\documentclass[twoside,11pt]{article}
\usepackage{jmlr2e}
\usepackage[utf8]{inputenc}
\usepackage{amsmath}
\usepackage{amsfonts}
\usepackage{mathtools}
\usepackage{url}
\usepackage{graphicx}
\usepackage{tikz}
\usepackage[vlined]{algorithm2e}
\usepackage{amssymb}
\usepackage{xcolor}
\usepackage{listings}
\usepackage{courier}
\usepackage{framed}
\usepackage{environ}
\usepackage{tcolorbox}
\usepackage{graphicx}
\usepackage{wrapfig}

 {\endMakeFramed}

\definecolor{ao}{rgb}{0.0, 0.5, 0.0}
\lstdefinestyle{Python}{
    language        = Python,
    basicstyle      = \ttfamily\footnotesize,
    keywordstyle    = \color{blue},
    keywordstyle    = [2]\color{teal}, 
    stringstyle     = \color{ao},
    commentstyle    = \color{red}\ttfamily,
    belowskip = 0.5em,
    aboveskip = 0.5em
}

\title{pygrank: A Python Package for Graph Node Ranking}

\ShortHeadings{Prior Editing for Graph Filter Posterior Fairness}{Krasanakis, Papadopoulos, Kompatsiaris and Symeonidis}
\firstpageno{1}

\begin{document}

\author{\name Emmanouil Krasanakis \email maniospas@iti.gr\\
       \name Symeon Papadopoulos \email papadop@iti.gr\\
       \name Ioannis Kompatsiaris \email ikom@iti.gr\\
       \addr 
       Centre for Research and Technology---Hellas
       \AND
       \name Andreas Symeonidis \email symeonid@ece.auth.gr\\
       \addr 
       Aristotle University of Thessaloniki
       }

\editor{This is an author preprint}

\maketitle

\begin{abstract}
    We introduce \textit{pygrank}, an open source Python package to define, run and evaluate node ranking algorithms. We provide object-oriented and extensively unit-tested algorithm components, such as graph filters, post-processors, measures, benchmarks and online tuning. Computations can be delegated to \textit{numpy}, \textit{tensorflow} or \textit{pytorch} backends and fit in back-propagation pipelines. Classes can be combined to define interoperable complex algorithms. Within the context of this paper we compare the 
    package with related alternatives and demonstrate its flexibility and ease of use with code examples.
\end{abstract}


\section{Introduction}
Representing data as graphs is popular in many domains; this involves designating data instances as nodes and linking those to related instances through edges. Various graph analysis approaches have been developed, with node ranking algorithms offering a powerful option when aiming to score nodes based on neighboring node scores. Typical node ranking techniques include spectral graph filters outlined by \cite{ortega2018graph}, also used in the seminal graph neural networks of \cite{klicpera2018predict} and \cite{huang2020combining}. However, most research approaches employ algorithms with little to no ablation studies, where different algorithms and parameters could better match the dynamics of analysed or---in case of deployed tools---new graphs.
\par
To support thorough investigation of node ranking algorithms and variations, we developed the \textit{pygrank} Python package. This implements a wide range of practices that define interoperable algorithms for machine learning and data mining pipelines.

\section{Source Code Organization}
\begin{wrapfigure}{r}{0.55\textwidth}
\centering
\includegraphics[width=0.53\textwidth,height=5.3cm]{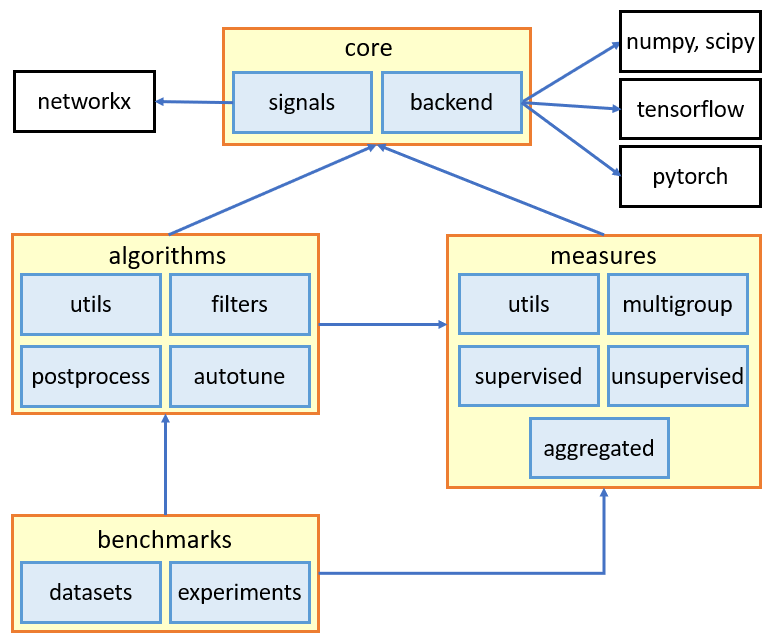}
\caption{Package architecture}\label{architecture}
\end{wrapfigure}

Figure~\ref{architecture} illustrates \textit{pygrank}'s functional components, along with the respective modules and dependencies. Running times of all components scale near-linearly with the number of edges, although multiplicative terms could apply due to iterative post-processing.
\vspace{0.3em}
~\\\textit{a) pygrank.core.} Manages backend and defines graph signals, which use backend primitives (e.g. numpy arrays, tensorflow tensors, pytorch tensors) to keep track of node scores.
\vspace{0.3em}
~\\\textit{b) pygrank.measures.} Provides supervised and unsupervised measures that compare graph signals or assess quantitative characteristics of the later in relation to graph dynamics.

~\\\textit{c) pygrank.algorithms.} Provides graph filters, post-processors and online tuning components, which can be combined to define a variety of machine learning algorithms. This includes fairness-aware post-processors outlined by \cite{krasanakis2020prioredit}. Base filters are wrapped so that all algorithms are used with the same method calls.
\vspace{0.3em}
~\\\textit{d) pygrank.benchmarks.} Helper methods to design node ranking experiments for collections of graphs, algorithms and training-test splits under evaluation measures.

\section{Comparison with Node Ranking alternatives}
A number of ad-hoc node ranking algorithms have been implemented in graph management packages, such as \textit{networkx} by \cite{hagberg2008exploring}, or deep graph learning packages, such as \textit{DGL} by \cite{wang2019deep}, \textit{pytorch geometric} by \cite{fey2019fast} and \textit{spektral} by \cite{grattarola2020graph}. Other solutions are implemented in scientific tools, such as \textit{textrank} by \cite{mihalcea2004textrank}. Finally, \textit{pygsp} by \cite{pygsp} provides many types of graph filters, but does not support non-spectral analysis and does not fit into machine learning pipelines.

Table~\ref{tab:comparison} compares packages that could run node ranking algorithms in new settings in terms of a) provision of \textit{ad-hoc} graph filters, b) supported \textit{backends}, c) ability to define \textit{general}-purpose filters, d) \textit{postpr}ocessing to improve outcomes, e) online \textit{tuning} and f) \textit{backpr}opagation support. We consider only base capabilities that pertain to node ranking algorithms and are usable by non-experts without additional development. For example, external autoML packages, such as \textit{autogluon} by \cite{erickson2020autogluon}, require more coding to use in deep graph learning setups. Overall, \textit{pygrank} introduces new functionality and combines advantages of other packages.

\begin{table}[htpb]
\footnotesize
    \centering
    \begin{tabular}{l c c c c c c}
        \textbf{package} & \textbf{ad-hoc} & \textbf{backends} & \rotatebox[origin=l]{0}{\textbf{general}} & \rotatebox[origin=l]{0}{\textbf{postpr.}} & \rotatebox[origin=l]{0}{\textbf{tuning}} & \rotatebox[origin=l]{0}{\textbf{backpr.}} \\
        \hline
         networkx & \checkmark & numpy \\
         pygsp & \checkmark & numpy & \checkmark\\
         dgl & \checkmark & mxnet, pytorch, tensorflow & & & & \checkmark\\
         pytorch geometric & \checkmark & pytorch & & & & \checkmark\\
         spektral & \checkmark & tensorflow & & & & \checkmark\\
         pygrank & \checkmark & numpy, pytorch, tensorflow & \checkmark  & \checkmark & \checkmark & \checkmark\\
    \end{tabular}
    \caption{Comparison of \textit{pygrank} with alternatives for node ranking algorithms.}
    \label{tab:comparison}
\end{table}

\section{Usage}

To demonstrate ease of use, in the following snippet we compare four ad-hoc graph filters to one tuned by an integrated variation of block coordinate descent. We use a common graph preprocessor to speed up execution and experiment on two datasets with community node labels to recommend new members by using half of the nodes for training and the rest for testing. Datasets are automatically downloaded and thus the code is ready to run after installing the package, e.g. with \textit{pip install pygrank}. Developers can provide local datasets, too. Results (could also be formatted as latex) assert the advantage of online tuning, which falls behind by at most 2\% compared to the best AUC.

\begin{lstlisting}[style=Python]
import pygrank as pg
datasets = ["EUCore", "Amazon"]
pre = pg.preprocessor(assume_immutability=True, normalization="symmetric")
algs = {"ppr.85": pg.PageRank(.85, preprocessor=pre),
        "ppr.99": pg.PageRank(.99, preprocessor=pre),
        "hk3": pg.HeatKernel(3, preprocessor=pre),
        "hk5": pg.HeatKernel(5, preprocessor=pre),
        "tuned": pg.ParameterTuner(preprocessor=pre)}
loader = pg.load_datasets_one_community(datasets)
pg.benchmark_print(pg.benchmark(algs, loader, pg.AUC, fraction_of_training=.5))
\end{lstlisting}
\begin{lstlisting}
               	 ppr.85  	 ppr.99  	 hk3  	 hk5  	 tuned 
EUCore         	 .86     	 .51     	 .90  	 .89  	 .88   
Amazon         	 .86     	 .92     	 .83  	 .87  	 .90   
\end{lstlisting}
One could add algorithm variations that employ the sweep post-processing of \cite{andersen2008local} to the investigation. Automatically generating such algorithms while adding a \textit{``+Sweep''} suffix to base names can be succeeded per:
\begin{lstlisting}[style=Python]
algs = algs | pg.create_variations(algs, {"+Sweep": pg.Sweep})
\end{lstlisting}

\textit{Pygrank} can also be used to define graph neural network architectures employing graph filters. For example, in the following code we implement the predict-then-propagate architecture of \cite{klicpera2018predict}, which defines a multilayer perceptron with the popular \textit{keras} package by \cite{chollet2015keras} and a node ranking algorithm with \textit{pygrank} to propagate perceptron prediction columns through the graph before passing these to a \textit{softmax} activation. We then load a graph with node features and create a 60-20-20 training-validation-test split of nodes. We train the model with a helper method \textit{gnn\_train} and load tensorflow's backend, so that its back-propagation passes through node ranking algorithms. Training would be the same if test labels were masked to zeros. Notably, graph filter propagation provided by our package takes up only two lines of code to define and run.

\begin{lstlisting}[style=Python]
import pygrank as pg
import tensorflow as tf

class APPNP:
    def __init__(self, num_inputs, num_outputs, hidden=64, dropout=0.5):
        self.mlp = tf.keras.Sequential([
                tf.keras.layers.Dropout(dropout, input_shape=(num_inputs,)),
                tf.keras.layers.Dense(hidden, activation=tf.nn.relu),
                tf.keras.layers.Dropout(dropout),
                tf.keras.layers.Dense(num_outputs, activation=tf.nn.relu),
            ])
        self.trainable_variables = self.mlp.trainable_variables
        self.ranker = pg.GenericGraphFilter([0.9]*10, renormalize=True, assume_immutability=True, tol=1.E-16)
    def __call__(self, graph, features, training=False):
        predict = self.mlp(features, training=training)
        propagate = self.ranker.propagate(graph, predict)
        return tf.nn.softmax(propagate, axis=1)
        
graph, features, labels = pg.import_feature_dataset('citeseer')
training, test = pg.split(list(range(len(graph))), 0.8)
training, validation = pg.split(training, 1-0.2/0.8)
pg.load_backend('tensorflow')  # explicitly load the appropriate backend
model = APPNP(features.shape[1], labels.shape[1])
pg.gnn_train(model, graph, features, labels, training, validation,
             optimizer = tf.optimizers.Adam(learning_rate=0.01),
             regularization = tf.keras.regularizers.L2(5.E-4))
print("Accuracy", pg.gnn_accuracy(labels, model(graph, features), test))
\end{lstlisting}

Even online tuning fits into machine learning pipelines. For example, the following code could be used in place of the APPNP class's ranker to automatically determine the node score diffusion parameter (\textit{params}[0]$\in[0.5,0.99]$) and the number of graph convolutions (\textit{params}$[1]\in[5,20]$). It also explicitly uses the \textit{numpy} backend for tuning which, as of writing, is several times faster than other backends in performing the sparse matrix multiplication implementing graph convolution.

\begin{lstlisting}[style=Python]
pre = pg.preprocessor(renormalize=True, assume_immutability=True)
self.ranker = pg.ParameterTuner(
    lambda params: pg.GenericGraphFilter([params[0]] * int(params[1]), preprocessor=pre, tol=1.E-16),
    max_vals=[0.99, 20], min_vals=[0.5, 5],
    measure=pg.KLDivergence, deviation_tol=0.1, tuning_backend="numpy")
\end{lstlisting}
'

\section{Conclusions}
In this document we presented \textit{pygrank}, a Python package for the definition and deployment of node ranking algorithms and demonstrated its practical usefulness. The package's source code and extensive documentation is publicly available under the Apache 2.0 license at the \textit{github} repository \url{https://github.com/MKLab-ITI/pygrank}. 
\par
Developers are encouraged to contribute to the code base. We follow continuous integration with github actions\footnote{\url{https://github.com/features/actions}} to make sure that testing passes and retains near-100\% coverage, as well as git hooks to automatically document new code components. The package has supported several publications in the domain of graph node ranking algorithms by \cite{krasanakis2020prioredit,krasanakis2020stopping,krasanakis2020unsupervised,krasanakis2020boosted}, whose algorithms have also been added to the code base.

\section*{Acknowledgements}
This work was partially funded by the European Commission under contract numbers H2020-951911 AI4Media and H2020-825585 HELIOS.
\bibliography{main}

\end{document}